\documentclass[journal]{IEEEtran}

\pdfoutput=1
\usepackage{times}
\usepackage{soul}
\usepackage{url}
\usepackage[draft]{hyperref}
\usepackage[utf8]{inputenc}
\usepackage[small]{caption}
\usepackage{graphicx}
\usepackage{subfigure}
\usepackage{amsmath}
\usepackage{booktabs}
\usepackage{algorithm}
\usepackage{algorithmic}
\usepackage{listings}
\usepackage{multirow}
\usepackage{psfig}
\begin{document}
%
\title{Convolutional Neural Network with Median Layers for Denoising Salt-and-Pepper Contaminations}
%
%
%

\author{\IEEEauthorblockN{Luming Liang\IEEEauthorrefmark{2},
        Sen Deng\IEEEauthorrefmark{3},
        Lionel Gueguen\IEEEauthorrefmark{4},
        Mingqiang Wei\IEEEauthorrefmark{3}\IEEEauthorrefmark{1},
        Xinming Wu\IEEEauthorrefmark{5},
        and~Jing~Qin\IEEEauthorrefmark{6}}
      \thanks{
      \IEEEauthorblockA{\IEEEauthorrefmark{2}L. Liang is with the Applied Science Group, Microsoft, Redmond, WA, 98052 USA e-mail: lulian@microsoft.com.}
      \IEEEauthorblockA{\IEEEauthorrefmark{3}S. Deng and M. Wei are with the School of Computer Science, Nanjing University of Aeronautics and Astronautics.}
      \IEEEauthorblockA{\IEEEauthorrefmark{4}L. Geuguen is with the Advanced Technology Group, Uber, CO, USA.}
      \IEEEauthorblockA{\IEEEauthorrefmark{5}X. Wu is with the University of Science and Technology of China.}
      \IEEEauthorblockA{\IEEEauthorrefmark{6}J. Qin is with the Hong Kong Polytechnic University.}
      Corresponding author: M. Wei (mqwei@nuaa.edu.cn).}}

\maketitle

\begin{abstract}
We propose a deep fully convolutional neural network with a new type of layer, named median layer, to restore images contaminated by the salt-and-pepper (s\&p) noise. A median layer simply performs median filtering on all feature channels. By adding this kind of layer into some widely used fully convolutional deep neural networks, we develop an end-to-end network that removes the extremely high-level s\&p noise without performing any non-trivial preprocessing tasks, which is different from all the existing literature in s\&p noise removal. Experiments show that inserting median layers into a simple fully-convolutional network with the $L_2$ loss significantly boosts the signal-to-noise ratio. Quantitative comparisons testify that our network outperforms the state-of-the-art methods with a limited amount of training data. The source code has been released for public evaluation and use (https://github.com/llmpass/medianDenoise).
\end{abstract}

\begin{IEEEkeywords}
Median layer, Deep neural network, Salt-and-pepper noise
\end{IEEEkeywords}

%
\IEEEpeerreviewmaketitle

\section{Introduction}

Image denoising is a well-studied yet not well-solved problem \cite{zhang2017beyond,zhang2018ffdnet,Liu2018,Fu2018Multimedia,Lehtinen2018,aaai_furuta_2019,chen2018DeepBoosting}, where the goal is to recover the underlying signal from its contaminated observation. The contaminations can be categorized into many different types according to their distributions and behaviors, e.g., additive (Gaussian) noise, shot (Poisson) noise, JPEG noise, etc. We focus on the \emph{salt-and-pepper} (s\&p) noise, which is an impulse contamination to the image. In an image with the s\&p noise, pixels become maximal or minimal values with a predefined probability, which is called the noise level, i.e. the higher this value is, the more pixels will be contaminated. The s\&p noise is a special case of random-value impulse noise defined in \cite{Fu2018Multimedia} and \cite{Lehtinen2018}. For a given noise level $p \in (0, 1)$, an s\&p contaminated image could be defined as
\begin{equation} \label{eq:spDefine}
I(i, j) = \begin{cases}
  0, & \text{$r_1 < p$ and $r_2 < 0.5$}\\
  255, &  \text{$r_1 < p$ and $r_2 \geq 0.5$}, \\
  I(i, j), & r_1 \geq p
  \end{cases}
\end{equation}
where both $r_1$ and $r_2$ are 2 random values generated on each pixel, with the former one determining if a pixel will be contaminated or not and later one controlling if that the pixel will turn to be the maximal (salt) value or the minimal (pepper) value.
From Equation \ref{eq:spDefine}, one observes that the s\&p noise is neither like the additive (Gaussian) noise, which can be fully separated from the signal \cite{zhang2017beyond}; nor like the shot (Poisson) noise, which is signal dependent \cite{Lehtinen2018}. It appears as the pure noise at the contaminated locations (called missing pixel in \cite{Lehtinen2018}) and therefore erases all signals there. This fact prevents us to use any optimization method in a continuous space to recover the signal, since the gradients estimated from missing pixels are totally not reliable and they will further scatter to other unpolluted locations. Traditional ways to recover images from s\&p pollution all require nonlinear searches and mappings. The search step \cite{Wang2011,Varghese2015,Julie2016} generally determines the locations of the contaminated pixels and the mapping step tries to give a feasible estimate at each contaminated pixel by weightedly averaging the similar pixel values around it. This set of filters are named switching filters. However, when the noise level is increasing, the search step becomes more and more unreliable. On the other hand, the signal estimation step will also be degraded by the high-level noise since the similarity estimation becomes intractable.

\begin{figure}
  \centering
  \setlength\tabcolsep{0.5pt}
  \begin{tabular}{cccc}
  \includegraphics[width=0.24\linewidth]{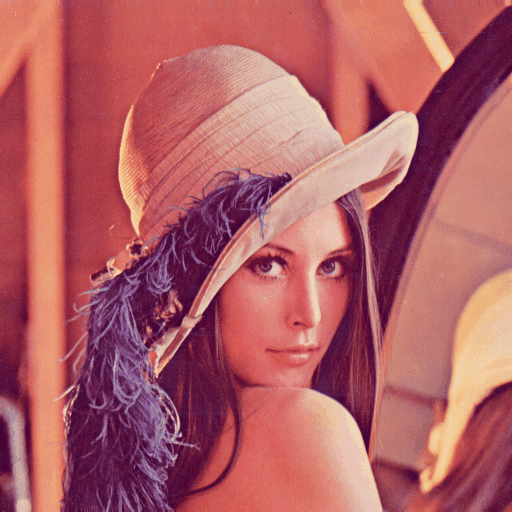} &
  \includegraphics[width=0.24\linewidth]{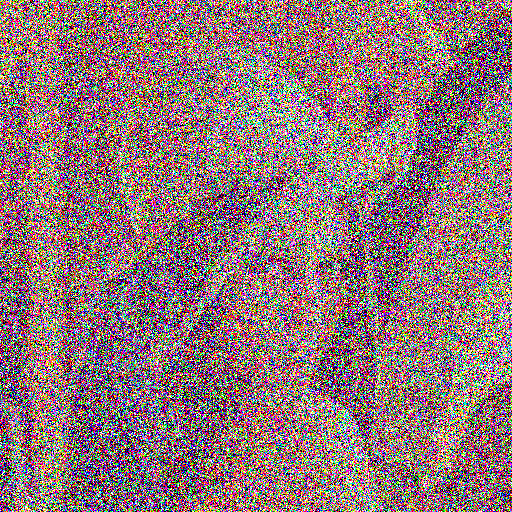} &
  \includegraphics[width=0.24\linewidth]{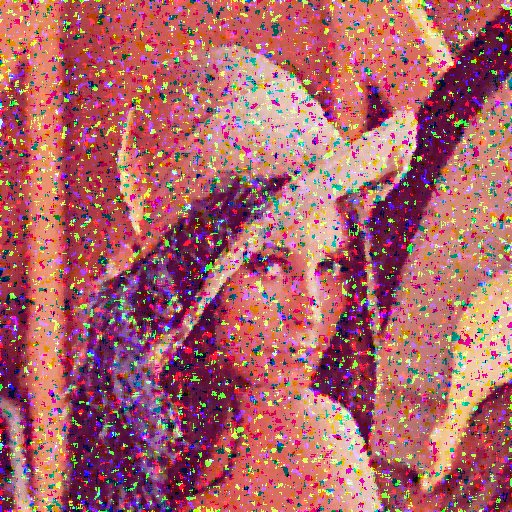} &
  \includegraphics[width=0.24\linewidth]{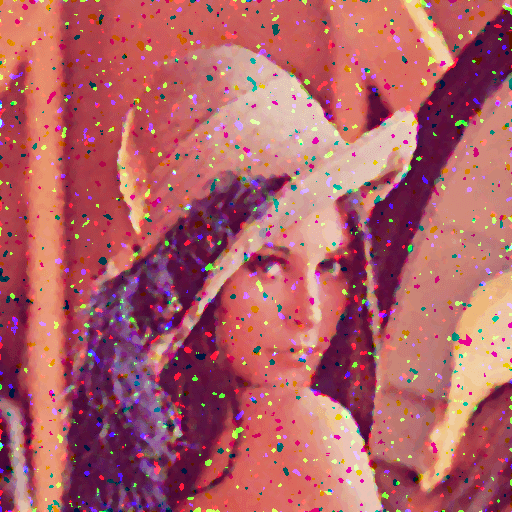}\\
  \small (a) Original &\small  (b) 70\% s\&p &\small  (c) Median5 &\small  (d) Median5 x2 \\
  \small  PSNR    &\small   6.72 db &\small 14.01 db &\small  19.14 db   \\
  \includegraphics[width=0.24\linewidth]{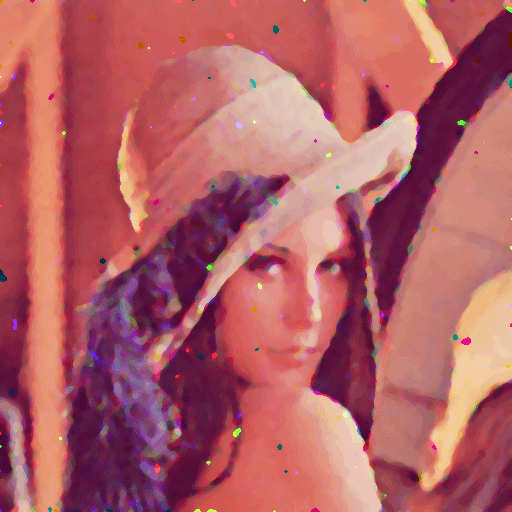} &
  \includegraphics[width=0.24\linewidth]{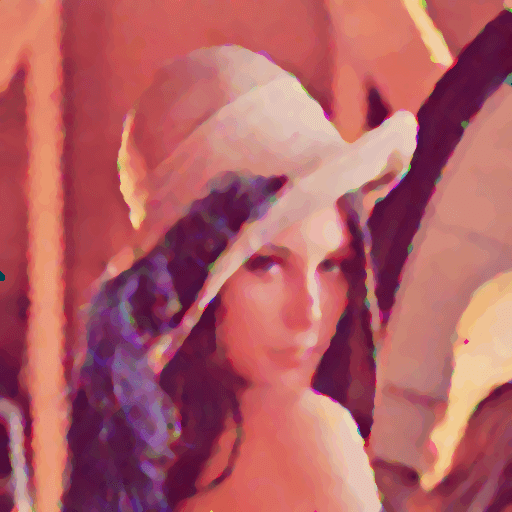} &
  \includegraphics[width=0.24\linewidth]{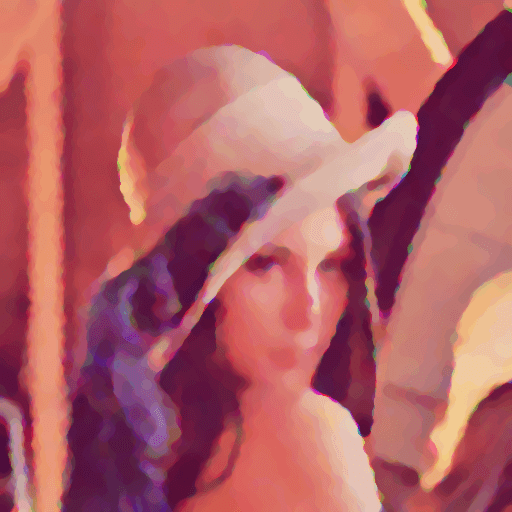} &
  \includegraphics[width=0.24\linewidth]{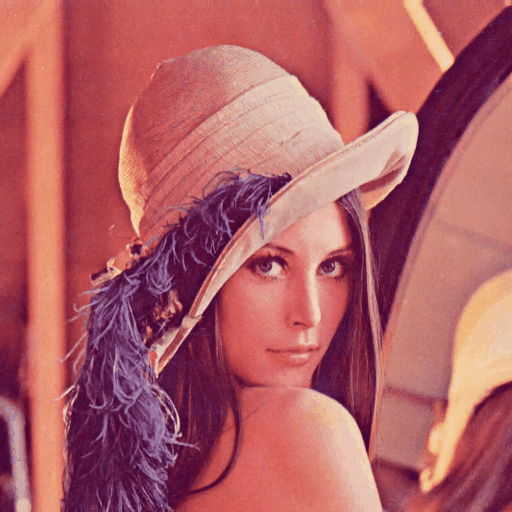} \\
  \small (e) Median5 x5 &\small  (f) Median5 x10 &\small  (g) Median5 x25 &\small  (h) Our method \\
  \small  24.09 db  &\small    24.89 db  &\small  24.52 db    &\small  33.07 db   \\
  \end{tabular}
  \caption{Classic Lenna image with the high-level s\&p noise contamination and filtering results with repeated median filter.
    }\label{fig:lenna}
\end{figure}

To alleviate this limitation, \cite{Fu2018Multimedia} uses switching templates to avoid noise disturbance in the process of measuring similarity. Based on the similarities, they extract repairable information in non-local regions instead of local patches. This filter is named as non-local switching filter (NLSF). The method uses a trained Convolutional Neural Network to finally refine the signal recovered by NLSF. Therefore, NLSF is considered as a prepocessing step to the neural network. This method is a combination of traditional methods and the learning-based method.

Besides the usual learning-based ideas that train models to denoise using pairs of clean images and their noisy versions, the noise-to-noise method \cite{Lehtinen2018} trains models only on noisy images. They discover that training without using clean images can achieve, sometimes even exceeds, the result obtained by training using ground truths. Following this paradigm, \cite{Lehtinen2018} shows the ability to remove the random-valued impulse noise, which can be considered as a superset of s\&p noise. To deal with the gradient loss problem introduced by pure noisy pixels, they adopt an annealed version of the ``$L_0$ loss" to replace the traditional $L_2$ loss. The loss function is gradually changing from $L_2$ to $L_0$ as the training progress. However, the speed of this annealing procedure must be carefully chosen (usually reducing the power of the norm on the loss function according to the number of iterations). When the prediction is far from the truth, the loss function must be closer to $L_2$; when the prediction is getting closer enough, $L_0$ becomes more favorable, since $L_0$ loss emphasis the number of different pixels, which leads the learning process to a detail amendment stage.

We introduce the use of local nonlinear search into the neural network without performing any pre-processing step and also avoid changing the loss function from $L_2$ loss to some other losses that are not easy to optimize. We resort to median filter \cite{Huang1979}, which is the first efficient method to denoise the salt-and-pepper noise. By incorporating the median-filter-like operations into deep neural networks, our method outperforms state-of-the-art methods. Details of our methodology as well as the model design can be found in Section \ref{sec:method}, evaluations are presented in Section \ref{sec:eval}. Section \ref{sec:conclusion} is for conclusion of our work. We release our source code, training dataset and pretrained models at \url{https://github.com/llmpass/medianDenoise} for reproducibility.

\section{Methodology} \label{sec:method}

\begin{figure}
  \centering
  \includegraphics[width=0.98\linewidth]{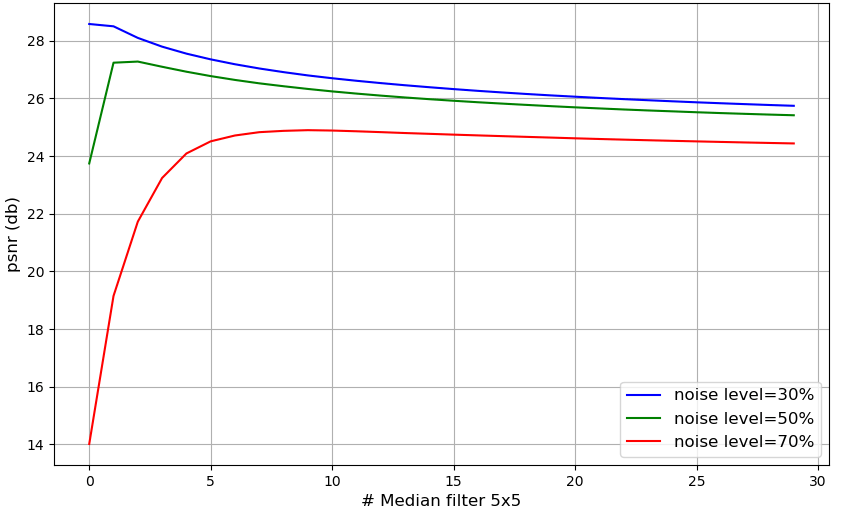}
  \caption{Peak Signal to Noise Ratio trends with respect to the number of iterations of repeated $5\times5$ median filters.
    }\label{fig:noise_psnr}
\end{figure}

Median filter is a traditional nonlinear filter which is especially efficient for removing impulse noise. It replaces the pixel centered in a given window with the median of this window.
As shown in Figure \ref{fig:lenna}, applying median filter on a highly contaminated image (b) removes spikes and therefore greatly improves the signal to noise ratio. Applying a 5$\times$5 median filter once (Figure \ref{fig:lenna}c) and twice (Figure \ref{fig:lenna}d),respectively, removes about $50\%$ and $90\%$ noise. A natural idea is to repeatedly apply the median filter upon the image until all spikes are replaced by the median in a fixed-size local window. It does remove the noise, however, it fails to recover the signal. The Peak Signal to Noise Ratio (PSNR) increases in the first several iterations but drops finally as the image becomes blocky and blurry, see Figure \ref{fig:lenna}g. This phenomena indicates that the median filter deviates the signal too much from its original shape, which is also the main reason why modern researchers abandon median filter in denoising s\&p noise.

In addition, the best PSNR value appears at different iterations of repeated median filtering when denoising different levels of noise. Figure \ref{fig:noise_psnr} shows the higher density the noise is, the more iterations of median filters are required.

\begin{figure}
  \centering
  \subfigure[Noisy signal]{\includegraphics[width=0.8\linewidth]{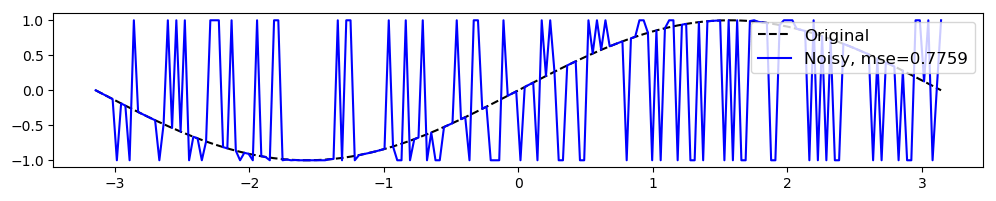}}
  \subfigure[Median filtered]{\includegraphics[width=0.8\linewidth]{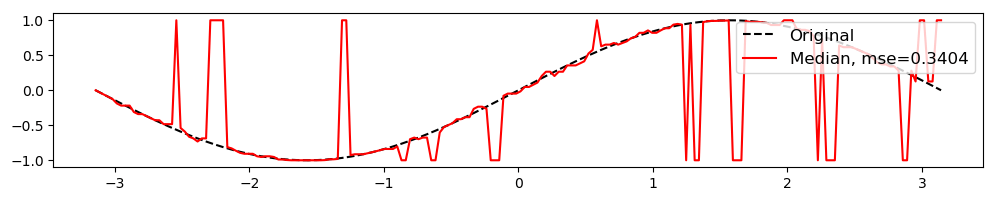}}
  \subfigure[Median filter applied twice]{\includegraphics[width=0.8\linewidth]{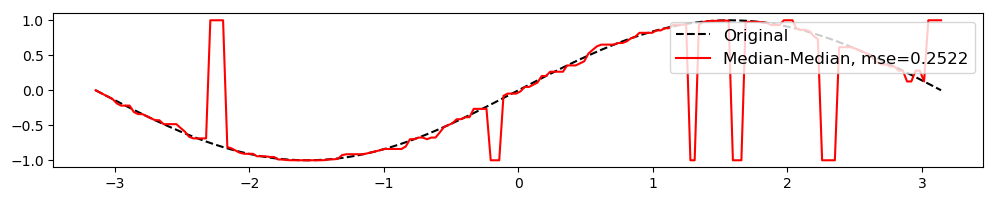}}
  \subfigure[Median filter applied 3 times]{\includegraphics[width=0.8\linewidth]{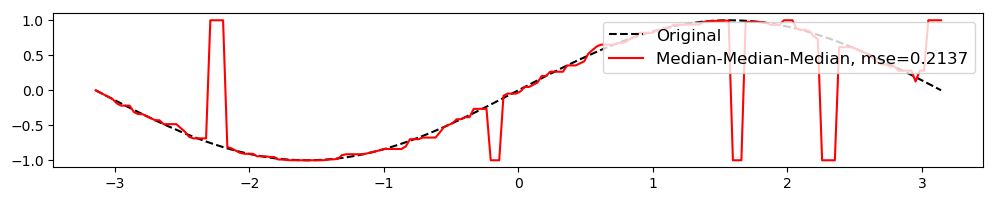}}
  \subfigure[Gaussian filtered]{\includegraphics[width=0.8\linewidth]{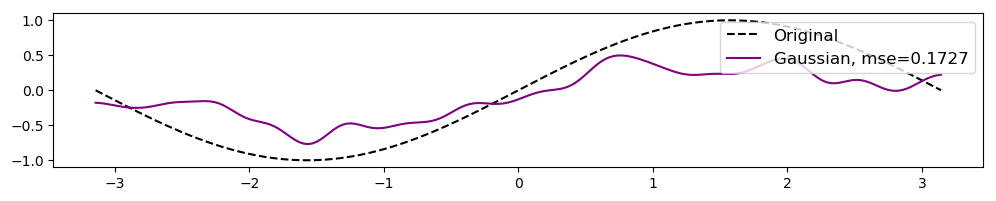}}
  \subfigure[Gaussian filter applied twice]{\includegraphics[width=0.8\linewidth]{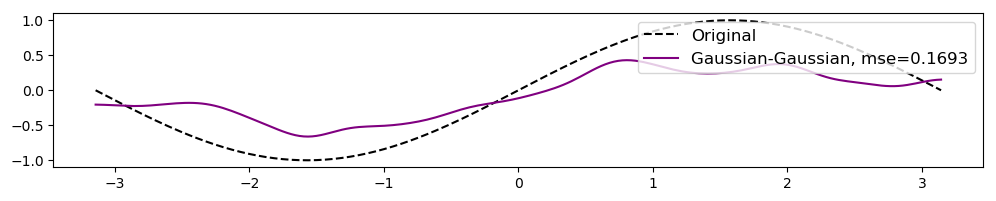}}
  \subfigure[Gaussian filter applied 3 times]{\includegraphics[width=0.8\linewidth]{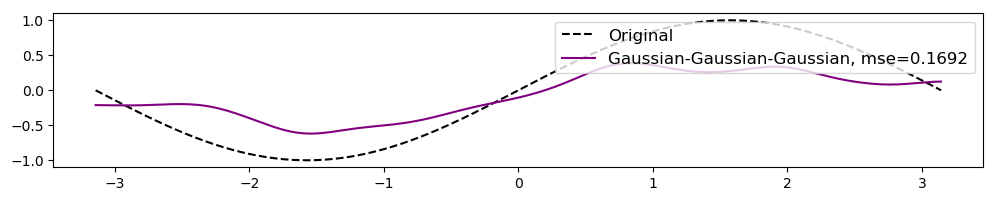}}
  \subfigure[Median filtered followed by Gaussian filtered]{\includegraphics[width=0.8\linewidth]{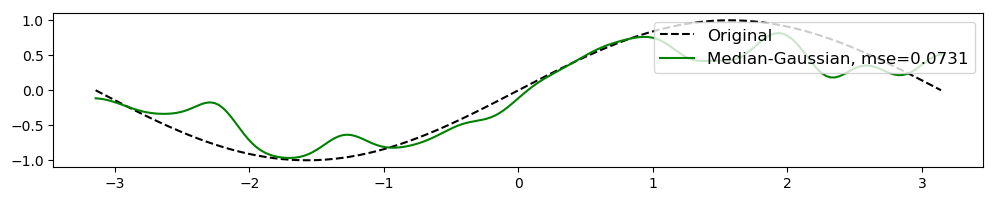}}
  \subfigure[filters applied: Median, Gaussian and Median]{\includegraphics[width=0.8\linewidth]{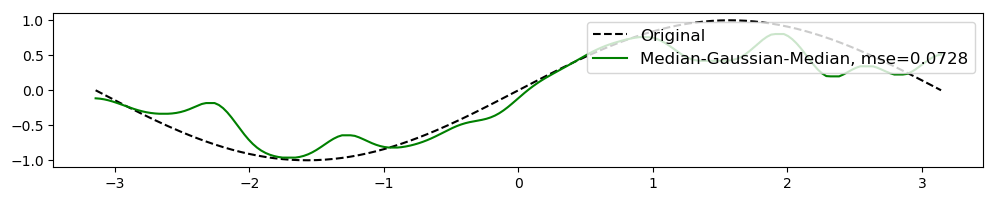}}
  \subfigure[filters applied: Median, Gaussian, Median and Gaussian]{\includegraphics[width=0.8\linewidth]{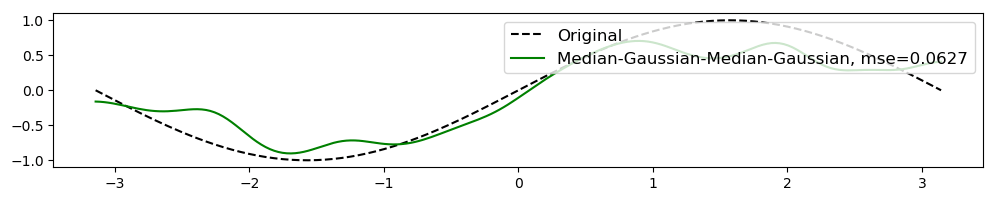}}
  \caption{1D signal denoising example using median filters and gaussian filters.
    }\label{fig:1dExample}
\end{figure}

Our basic idea is to keep the ability of spike removal from the traditional median filter but try to recover the degradations introduced by it. Figure \ref{fig:1dExample} illustrates a simple 1D synthetic example. We contaminate an evenly-sampled 1D sine function (dotted curves in Figure \ref{fig:1dExample}a) by $50\%$-level s\&p noise. After that, we tried to use different ways to recover the clean signal:
\begin{enumerate}
  \item Repeated Median filters, see Figure \ref{fig:1dExample}b-d;
  \item Repeated Gaussian filters, see Figure \ref{fig:1dExample}e-f;
  \item Alternating Median and Gaussian filters, see Figure \ref{fig:1dExample}h-j.
\end{enumerate}
Here, all Median and Gaussian filters have the same window size that equals to 5 pixels. One may observe the third schema yields the best approximations (green curves) to the original sine function, no matter in the aspect of the signal shape or mean square errors (mse) between the smoothed curves (solid) and the true signal (dotted) curve. Using only Median filters creates plateau-like artifacts; using only Gaussian filters over smooth the noisy curves. By alternating Median and Gaussian filters, apparent plateau-like artifacts are washed out while the resulting curve still stays close to the true signal. The quick-dropping mse values between smoothed curves generated by the third schema and the truth quantitatively support our observation.

We leverage these observations to design our deep neural network model for 2d image denoising. We replace the Gaussian filter, which is a fix-parameter smoothing filter, by a set of learnable convolution operations and thus design an end-to-end fully convolutional network with Median and other convolutions alternatingly appearing.

Instead of directly applying median filters on the images, we implement median filtering as a neural network operation and perform it on different feature channels. In this way, we essentially remove spikes in different feature spaces and then combine the de-spiked features to predict a better noise removed image. On one hand, the median filtering in the feature space acts just like the switch filters in the traditional methods \cite{Wang2011,Fu2018Multimedia}; on the other hand, the de-spike ability introduced by median operations allow the gradients to pass through the non-noisy pixels.

\subsection{Median layer definition}

Median filter is applied to each element of a feature channel in a moving window fashion. For example, an input image that consists of RGB channels, corresponds to 3 feature channels; a set of features generated after the convolution generally contains many number of channels. For each feature channel, we first extract a set of given size ($3\times3$, $5\times5, ...)$ size patches centered at each pixel. Then, we find the median of the sequence formed by all elements in that patch. We show a simple tensorflow/python implementation of this median filter layer in Listing \ref{list:median}. Here, parameter $x$ is a channel of the input tensor and $k$ denotes an integer kernel size.

\begin{lstlisting}[language=Python, basicstyle=\tiny, label={list:median}, caption={Tensorflow implementation of Median layer},captionpos=b]
def find_medians(x, k=3):
    patches = tf.extract_image_patches(
            x,
            ksizes=[1, k, k, 1],
            strides = [1, 1, 1, 1],
            rates=[1, 1, 1, 1],
            padding='SAME')
    m_idx = int(k*k/2 + 1)
    top, _ = tf.nn.top_k(patches, m_idx, sorted=True)
    median = tf.slice(top, [0, 0, 0, m_idx-1], [-1, -1, -1, 1])
    return media
\end{lstlisting}

In practice, this median layer is applied on each feature channel and then we concatenate them to form a new set of features, e.g. median layer will be applied 64 times given a set of $64$ feature channels generated by Convolutions.

\subsection{Network architecture}

\begin{figure*}[!ht]
  \centering
  \subfigure[Fully convolutional network with median layers in between residual blocks.]{\includegraphics[width=0.7\linewidth]{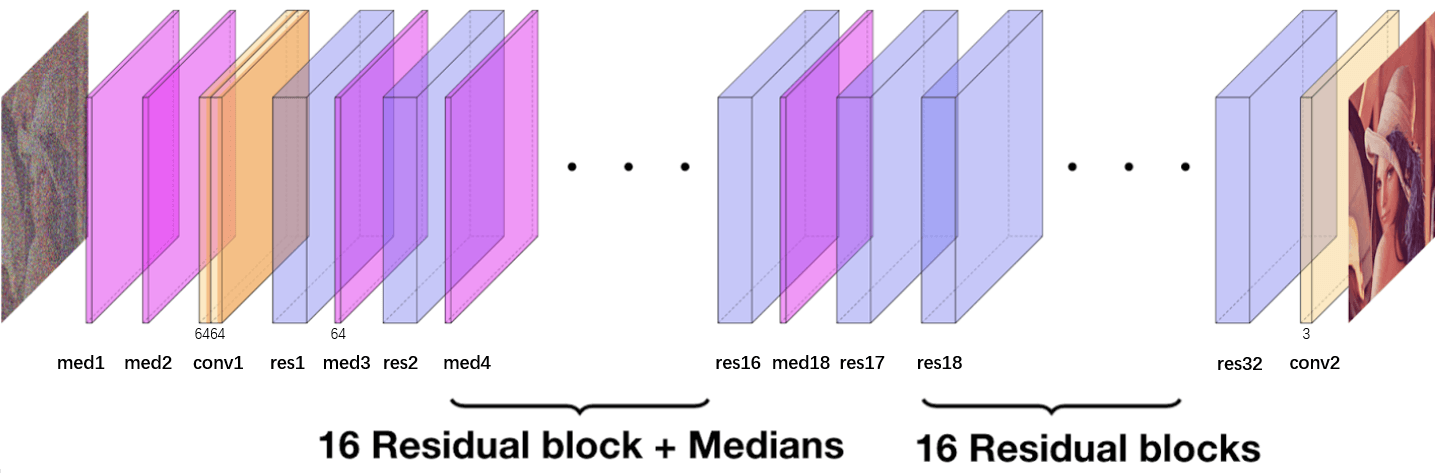}}
  \subfigure[Our residual blocks.]{\includegraphics[width=0.28\linewidth]{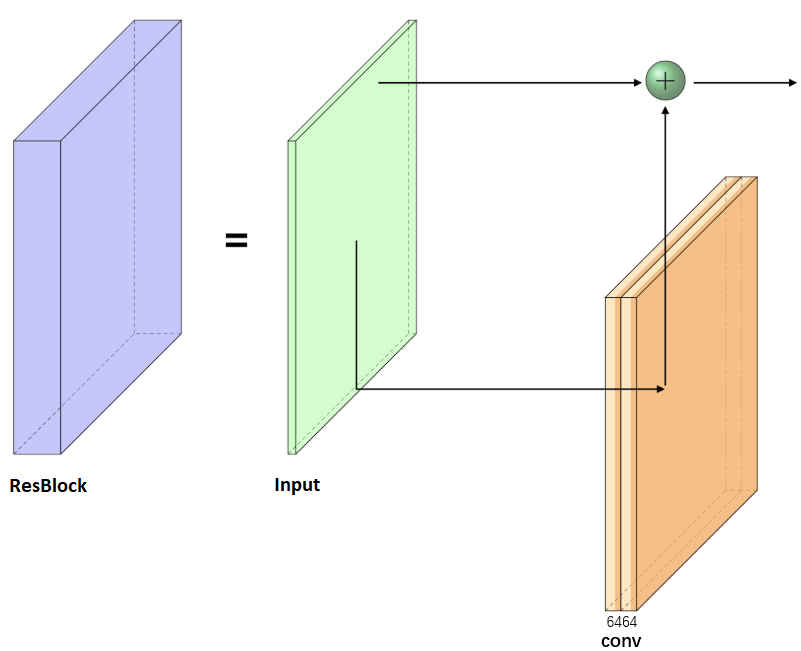}}
  \caption{Our network structure.
    }\label{fig:network}
\end{figure*}

As shown in Figure \ref{fig:network}a, our network is a fully convolutional network, so that no restrictions are posed on the size of the input.
It starts with 2 consecutive median layers, which are then followed by a sequence of residual blocks and median layers. The last part of the network is just residual blocks without inserting median layers in between them. In practice, we only insert median layers into the first half of the sequence of residual blocks. The first part of the network is dedicated to remove noise from the image, the second half of the network is designed for recovering the signal.

We choose to generate $64$ features per convolution layer and our residual block is designed as a skip connection over 2 $64$-convolutions, followed by batch normalization layers and nonlinear activations (relu in practice), as shown in Figure \ref{fig:network}b.

As mentioned beforehand, we stick to use the simplest $L_2$ loss as our objective function. This loss is simply defined as the the mean square error of the estimated image and the ground truth image, as minimizing mse directly relates to increase of denoise metrics psnr. Details can be found in Equation \ref{eq:mse}.

\begin{table*}
\centering
\begin{tabular}{lrrrrrr}
\toprule
noise level  & ConvRelu 16    & ConvRelu 16  & ResBlock 16    & ResBlock 16  & ResBlock 32    & ResBlock 32 \\
             &   \textbf{without} median  & \textbf{with} median &   \textbf{without} median  & \textbf{with} median &   \textbf{without} median  & \textbf{with} median \\
\midrule
30\%       & 31.15  & \textbf{33.82}    & 40.38  & \textbf{40.89}   & 36.86 & \textbf{40.90} \\
50\%       & 30.15  & \textbf{32.01}    & 36.55  & \textbf{36.93}   & 36.86 & \textbf{37.28} \\
70\%       & 28.88  & \textbf{29.37}    & 31.98  & \textbf{32.23}   & 32.22 & \textbf{32.40} \\
\bottomrule
\end{tabular}
\caption{PSNR (db) comparisons w/o Median layers on BSD300.}
\label{tab:median_effect_table}
\end{table*}

\section{Evaluation} \label{sec:eval}

We design several experiments to evaluate the properties of median layers (Section \ref{sec:effectMedian}) and performances of the proposed network (Section \ref{sec:compare}).

\subsection{Training and testing setup}

For fair comparisons, we train all models with the same data set described in \cite{Dong2015PAMI} that contains $91$ different images, which is also employed in other works \cite{Fu2018Multimedia}. Since our network is a fully convolutional network, the input size can be arbitrary. We first resize these $91$ images to $200\times200$ and then we generate $70\times70$ patches from them as clean images. We degrade each patch by the s\&p noise with levels from $10\%$ to $90\%$ with a step equals to $10\%$ as a sequence of noisy images. The models are trained to learn a series of weights in layers that can transfer the input noisy image to the clean image.

To quantitatively compare the performance of different methods, we perform denoising on 3 sets of the images. The first set of image consist of some classical images in the image processing field (also used in \cite{Fu2018Multimedia}); the second set is BSD300 \cite{MartinFTM01} (\url{https://www2.eecs.berkeley.edu/Research/Projects/CS/vision/grouping/segbench/BSDS300/html/dataset/images.html}); the third set is Kodak Image Dataset (\url{http://r0k.us/graphics/kodak/}), which has been widely used as the evaluation set \cite{zhang2017beyond,zhang2018ffdnet,Lehtinen2018}.

The metric considered in the comparison is Peak Signal to Noise Ratio (PSNR). It is defined by
\begin{equation}
PSNR = 10\log_{10}(\frac{255^2}{MSE}), \label{eq:psnr}
\end{equation}
where $MSE$ is the mean-squared error between two $M \times N$ 8-bit images $I_1$ and $I_2$, defined by
\begin{equation}
MSE = \dfrac{\sum_{M,N}[I_1(m,n) - I_2(m,n)]^2}{M \times N}. \label{eq:mse}
\end{equation}

\subsection{Effects of the median layer} \label{sec:effectMedian}

The first experiment is designed to show the effectiveness of the median layer. We trained several pairs of fully convolutional networks mainly consisting of residual or convolution-batchNorm-relu blocks, but one with median layers and one without them.

\begin{figure}
  \centering
  \setlength\tabcolsep{0.05pt}
  \begin{tabular}{cc}
  \includegraphics[width=0.45\linewidth]{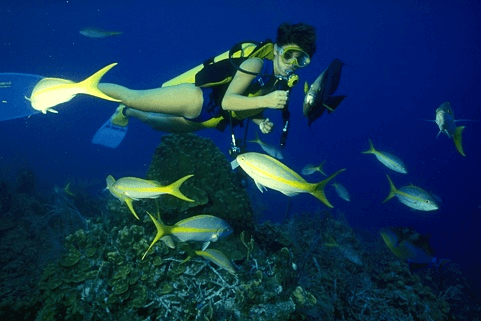} &
  \includegraphics[width=0.45\linewidth]{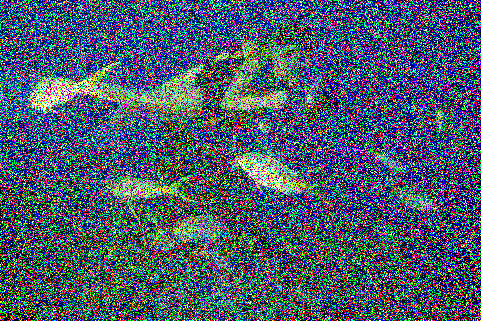} \\
  \small (a) Original &\small  (b) 70\% s\&p (6.72 db) \\
  \includegraphics[width=0.45\linewidth]{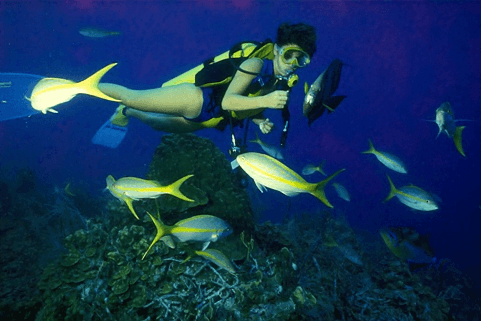} &
  \includegraphics[width=0.45\linewidth]{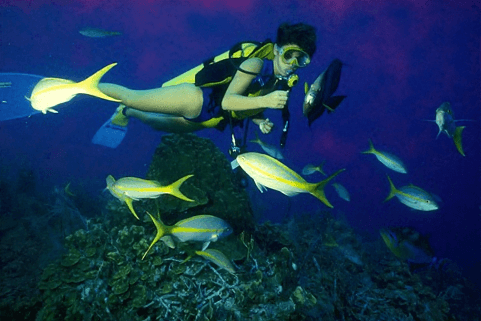}\\
  \small (c) W/ medians (32.16db) &\small  (d) W/O medians (28.70db) \\
  \end{tabular}
  \caption{Denoise results with and without median layers on an image of BSD300, measured by PSNR (db).
    }\label{fig:WOmedians}
\end{figure}

\begin{figure}
  \centering
  \includegraphics[width=0.8\linewidth]{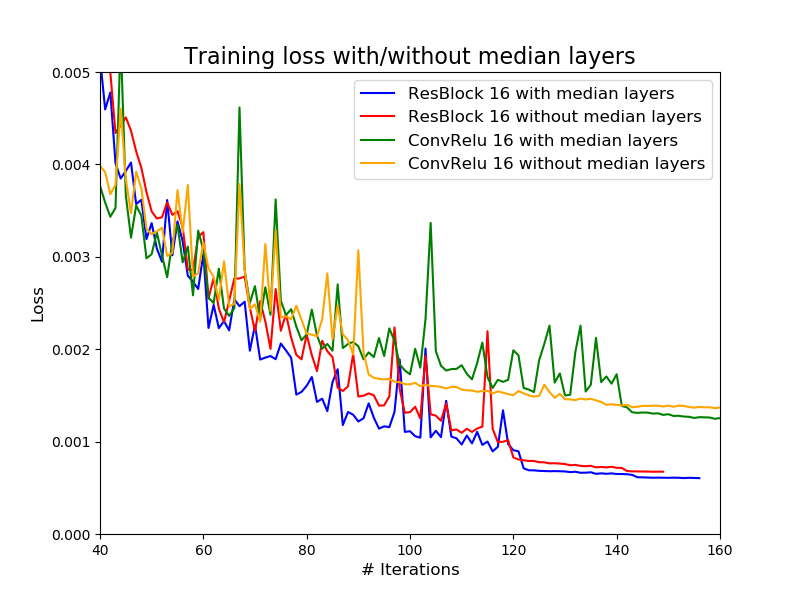}
  \caption{Training losses with and without median layers.
    }\label{fig:median_loss}
\end{figure}

\begin{table*}
\centering
\begin{tabular}{l|ccccccccc}
\toprule
Image & Noise level &  DBA \cite{Srinivasan2007SPL}  & NASNLM \cite{Varghese2015} & PARIGI \cite{Julie2016} & NLSF \cite{Fu2018Multimedia} & NLSF-MLP \cite{Burger2012} & NLSF-CNN \cite{Fu2018Multimedia} & Noise2Noise \cite{Lehtinen2018} & Ours \\
\midrule
\multirow{3}{*}{\includegraphics[width=0.05\linewidth]{lenna.png}}
 & 30\% & 34.42 & 28.09 & 33.90 &34.20 & 30.80 & 35.38 & 36.39& \textbf{37.04} \\
 & 50\% & 30.11 & 26.15 & 29.91 & 30.12 & 29.28 & 32.55 & 34.68 & \textbf{35.00} \\
 & 70\% & 25.84 & 25.97 & 25.22 & 25.79 & 27.63 & 30.18 & 32.83 & \textbf{33.07} \\
\midrule
\multirow{3}{*}{\includegraphics[width=0.05\linewidth]{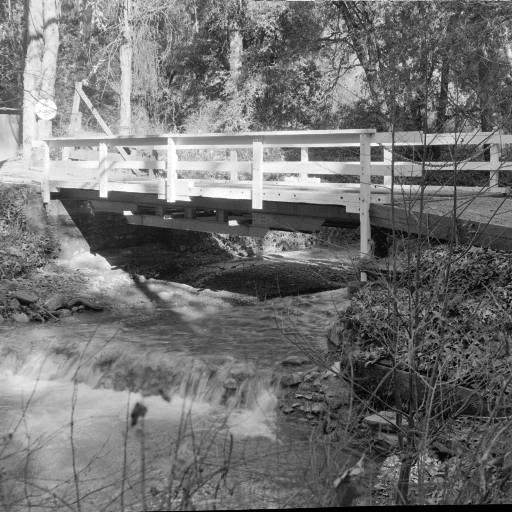}}
 & 30\% & 28.07 & 23.68 & 25.19 & 28.21 & 25.19 & 28.71 & 30.89 & \textbf{40.46} \\
 & 50\% & 24.24 & 22.91 & 22.61 & 24.45 & 23.86 & 26.01 & 27.96 & \textbf{34.83} \\
 & 70\% & 21.12 & 22.63 & 20.06 & 21.02 & 22.61 & 24.11 & 25.09 & \textbf{29.96} \\
\midrule
\multirow{3}{*}{\includegraphics[width=0.05\linewidth]{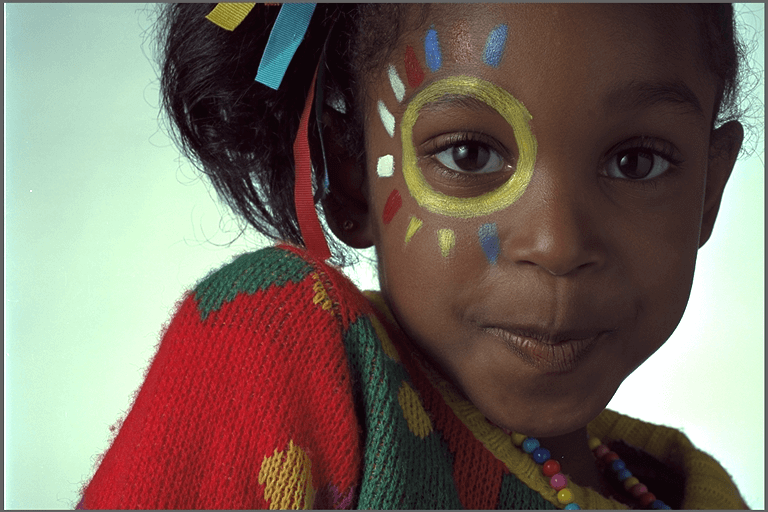}}
 & 30\% & 29.41 & 20.61 & 29.74 & 32.88 & 29.64 & 33.47 & 39.98 & \textbf{40.65} \\
 & 50\% & 27.47 & 16.69 & 27.25 & 29.66 & 28.28 & 30.92 & 36.13 & \textbf{38.84} \\
 & 70\% & 24.99 & 16.32 & 24.29 & 26.33 & 26.90 & 29.06 & 30.55 & \textbf{33.29} \\
\midrule
\multirow{3}{*}{\includegraphics[width=0.05\linewidth]{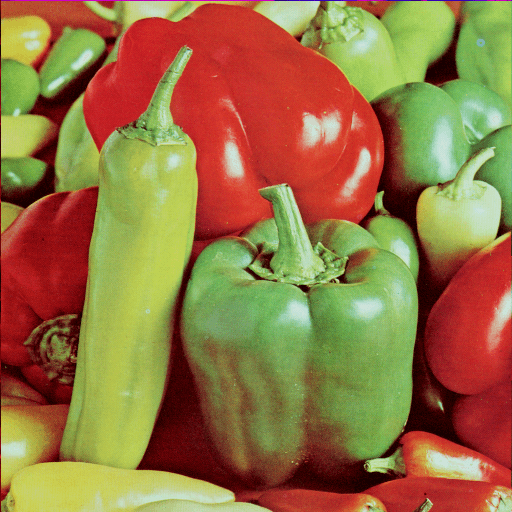}}
 & 30\% & 26.85 & 22.38 & 28.88 & 32.27 & 30.01 & \textbf{32.99} & 30.70 & 30.83 \\
 & 50\% & 25.27 & 21.82 & 25.44 & 27.99 & 28.57 & \textbf{30.23} & 29.86 & 30.07 \\
 & 70\% & 22.11 & 21.58 & 21.46 & 23.04 & 27.04 & 27.70 & 28.79 & \textbf{29.05} \\
\midrule
BSD300 & 30\% & 29.92 & 25.74 & 12.04 & 30.01 & 29.77 & 30.87 & 39.83 & \textbf{40.90} \\
\multirow{2}{*}{average} & 50\% & 26.32 & 24.50 &  6.01 & 26.25 & 26.19 & 27.84 & 35.92 & \textbf{37.28} \\
 & 70\% & 22.81 & 24.65 & 5.42 & 22.85 & 26.19 & 25.35 & 31.42 & \textbf{32.40} \\
\bottomrule
\end{tabular}
\caption{PSNR (db) Comparisons with state-of-the-arts on a set of classic images and BSD300 image database. Best performances of every noise levels of different images are in \textbf{bold}.}
\label{tab:pnsr_classic}
\end{table*}

\begin{table}
\centering
\begin{tabular}{l|ccc}
\toprule
Noise level & DeepBoosting \cite{chen2018DeepBoosting} & Noise2Noise \cite{Lehtinen2018} & Ours \\
\midrule
30\% & 21.69 & 34.95 & \textbf{36.39} \\
50\% & 19.50 & 32.27 & \textbf{34.35} \\
70\% & 15.74 & 30.49 & \textbf{31.56} \\
\bottomrule
\end{tabular}
\caption{PSNR (db) Comparisons with state-of-the-arts on Kodak image database. Best performances of every noise levels of different images are in \textbf{bold}.}
\label{tab:pnsr_kodak}
\end{table}

\begin{figure*}
  \centering
  \small
  \setlength\tabcolsep{0.05pt}
  \begin{tabular}{cccccccc}
  \multicolumn{2}{c}{\includegraphics[width=0.25\linewidth]{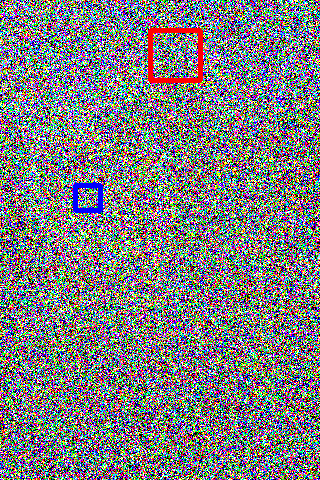}} &
  \multicolumn{2}{c}{\includegraphics[width=0.25\linewidth]{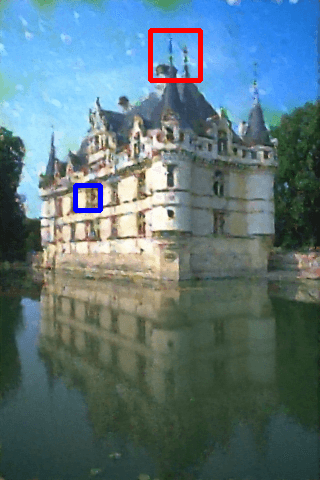}} &
  \multicolumn{2}{c}{\includegraphics[width=0.25\linewidth]{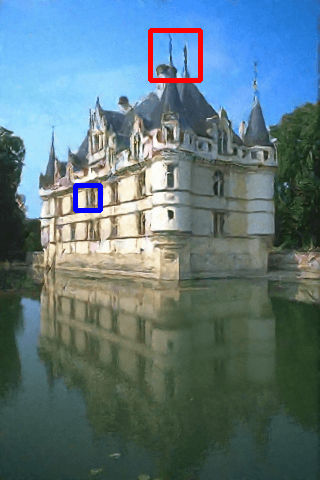}} &
  \multicolumn{2}{c}{\includegraphics[width=0.25\linewidth]{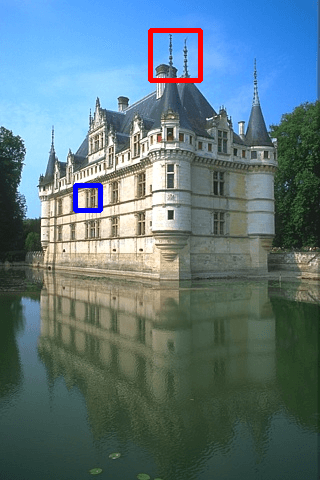}} \\
  \includegraphics[width=0.125\linewidth]{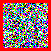} &
  \includegraphics[width=0.125\linewidth]{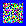} &
  \includegraphics[width=0.125\linewidth]{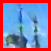} &
  \includegraphics[width=0.125\linewidth]{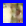} &
  \includegraphics[width=0.125\linewidth]{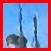} &
  \includegraphics[width=0.125\linewidth]{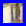} &
  \includegraphics[width=0.125\linewidth]{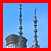} &
  \includegraphics[width=0.125\linewidth]{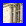} \\
  \multicolumn{2}{c}{(a)} &
  \multicolumn{2}{c}{(b)} &
  \multicolumn{2}{c}{(c)} &
  \multicolumn{2}{c}{(d)} \\
  \multicolumn{2}{c}{\includegraphics[width=0.25\linewidth]{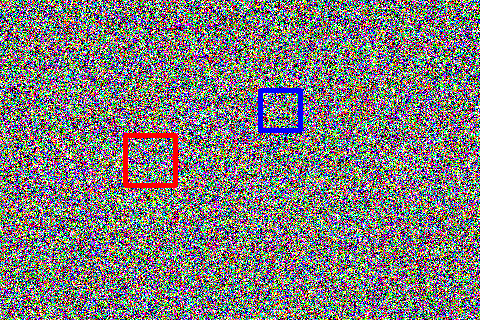}} &
  \multicolumn{2}{c}{\includegraphics[width=0.25\linewidth]{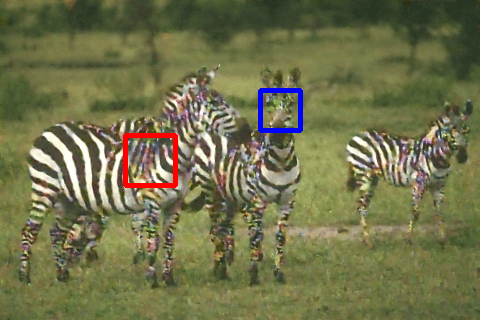}} &
  \multicolumn{2}{c}{\includegraphics[width=0.25\linewidth]{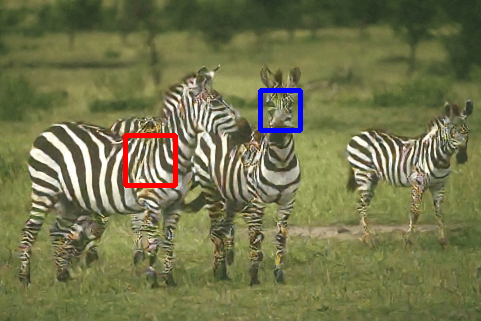}} &
  \multicolumn{2}{c}{\includegraphics[width=0.25\linewidth]{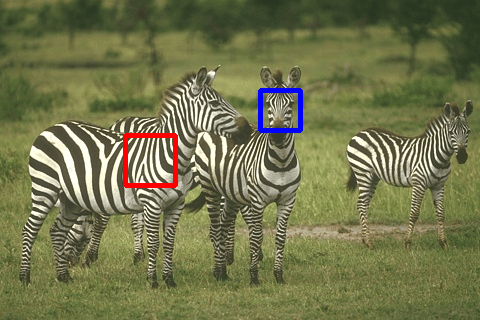}} \\
  \includegraphics[width=0.125\linewidth]{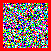} &
  \includegraphics[width=0.125\linewidth]{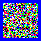} &
  \includegraphics[width=0.125\linewidth]{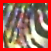} &
  \includegraphics[width=0.125\linewidth]{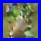} &
  \includegraphics[width=0.125\linewidth]{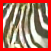} &
  \includegraphics[width=0.125\linewidth]{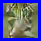} &
  \includegraphics[width=0.125\linewidth]{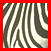} &
  \includegraphics[width=0.125\linewidth]{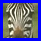} \\
  \multicolumn{2}{c}{(e)} &
  \multicolumn{2}{c}{(f)} &
  \multicolumn{2}{c}{(g)} &
  \multicolumn{2}{c}{(h)} \\
  \multicolumn{2}{c}{\includegraphics[width=0.25\linewidth]{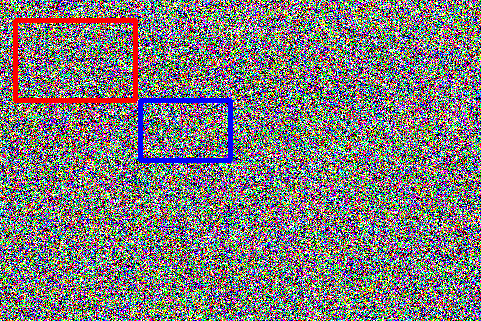}} &
  \multicolumn{2}{c}{\includegraphics[width=0.25\linewidth]{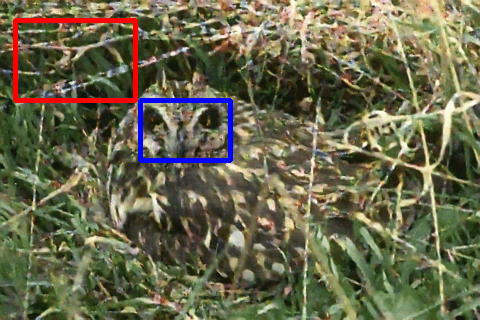}} &
  \multicolumn{2}{c}{\includegraphics[width=0.25\linewidth]{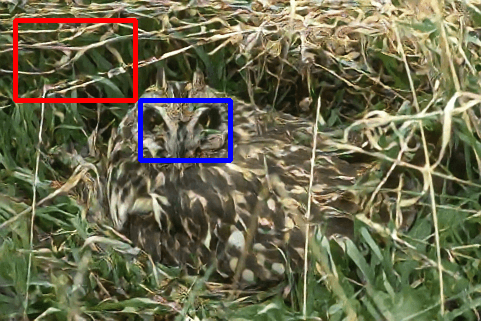}} &
  \multicolumn{2}{c}{\includegraphics[width=0.25\linewidth]{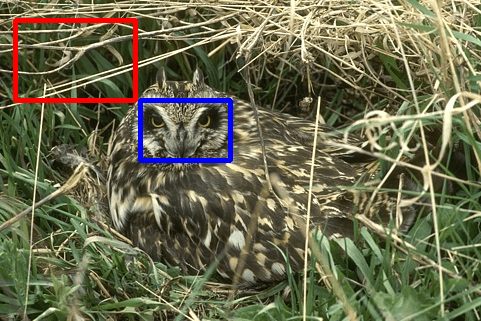}} \\
  \includegraphics[width=0.125\linewidth]{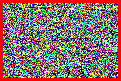} &
  \includegraphics[width=0.125\linewidth]{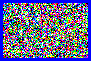} &
  \includegraphics[width=0.125\linewidth]{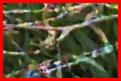} &
  \includegraphics[width=0.125\linewidth]{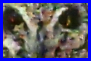} &
  \includegraphics[width=0.125\linewidth]{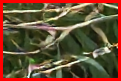} &
  \includegraphics[width=0.125\linewidth]{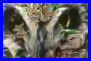} &
  \includegraphics[width=0.125\linewidth]{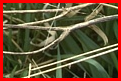} &
  \includegraphics[width=0.125\linewidth]{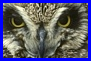} \\
  \multicolumn{2}{c}{(i)} &
  \multicolumn{2}{c}{(j)} &
  \multicolumn{2}{c}{(k)} &
  \multicolumn{2}{c}{(l)} \\
  \multicolumn{2}{c}{\small 90\% s\&p } &
  \multicolumn{2}{c}{\small noise2noise} &
  \multicolumn{2}{c}{\small ours} &
  \multicolumn{2}{c}{\small ground truth} \\
  \end{tabular}
  \caption{Detailed comparisons between noise2noise and our model. Our model outperforms noise2noise consistently on different challenges: 1) smoothly changing background (the first row); 2) white and black strips (the second row) and 3) noise-like natural scene.} \label{fig:detail}
\end{figure*}

We train two sets of deep fully convolutional networks, the first set of networks are traditional ones that do not contain any median layers, the second set of networks are the counterparts of the first set with median layers inserted into them with the same strategy shown in Figure \ref{fig:network}a, i.e. the first half of the network contains median layers, the second half does not. The networks in the first set consist of repeated blocks of convolution, batch normalization and activation or repeated residual blocks as shown in Figure \ref{fig:network}b.

Losses in Figure \ref{fig:median_loss} shows how median layers boost the PSNR value of the network. Training losses of two networks with median layers inserted converge to a better minima comparing to the losses without the median layers. The ``ConvRelu 16" network in Figure \ref{fig:median_loss} is a \textbf{DnCnn} \cite{zhang2017beyond} style network, which consists of 16 stacked Convolution-BatchNormalization-Activation units. The ``ResBlock 16" network in Figure \ref{fig:median_loss} is formed by simply replacing the Convolution-BatchNormalization-Activation units to residual blocks shown in Figure \ref{fig:network}b. All convolution layers here generate 64 features.

PSNR comparisons of models with and without median layers inserted (Table \ref{tab:median_effect_table}) show the improvements of PSNR. The PSNR values of models with median layers are usually $0.5$db higher than the ones that do not have them.

\subsection{Comparisons to the state-of-the-arts} \label{sec:compare}

Both quantitative and qualitative comparisons to the state-of-the-arts are performed in this section.

\subsubsection{Quantitative comparisons} \label{sec:quantCompare}

We quantitatively compare our network in Figure \ref{fig:network}a to several state-of-the-art methods. Baselines include 5 traditional methods: Decision-Based
Algorithm (DBA) \cite{Srinivasan2007SPL}, Adaptive Switching Non-local Filter (NASNLM) \cite{Varghese2015}, PARIGI \cite{Julie2016}, NLSF \cite{Fu2018Multimedia} (prepocessing part of NLSF-CNN), NLSF-MLP (NLSF with multi-layer perception proposed in \cite{Burger2012}) and 2 most recent neural network based methods: NLSF-CNN \cite{Fu2018Multimedia} and Noise2Noise \cite{Lehtinen2018}, as shown in Table \ref{tab:pnsr_classic}. Many methods here are designed for denoising s\&p noise with moderate levels, therefore, we choose to evaluate the methods under noise levels equal to $30\%$, $50\%$ and $70\%$.

In addition, we also compare our method to DeepBoosting \cite{chen2018DeepBoosting} and Noise2Noise \cite{Lehtinen2018} on Kodak image dataset, as shown in Table \ref{tab:pnsr_kodak}.

Our method outperforms most of the state-of-the-arts besides the pepper image. Comparing to current best baseline method Noise2Noise \cite{Lehtinen2018}, PSNR values achieved by our model is about 1-2db higher in average and the severer the noise contamination, the comparably better our method performs.

\subsubsection{Qualitative comparisons on extremely high-level noise}

We further qualitatively compare our method to noise2noise method \cite{Lehtinen2018} on denoising extremely high-level s\&p noise (noise level equals to $90\%$). In Figure \ref{fig:detail}, we choose three images from BSD300 dataset, where different challenges can be found there:
\begin{itemize}
  \item both sharp feature and smooth background exist in the first image (Figure \ref{fig:detail}d);
  \item pure black and white interphase pattern in the second image (Figure \ref{fig:detail}h);
  \item noise-like nature scene background (Figure \ref{fig:detail}l).
\end{itemize}
The left-most column in Figure \ref{fig:detail} shows the contaminated images, which are the noisy version of their counterparts in the right-most column. One may hardly see the contours of the original salient objects there, since $90\%$ of pixels become either maximal or minimal values.

Our method performs consistently better than noise2noise on all of these challenges. In Figure \ref{fig:detail}b, noise2noise generates many small white blocky artifacts on the sky (red rectangle) and also blurs the sharp edges (blue rectangle) of the windows. Both of these 2 degradations are alleviated in our result shown in Figure \ref{fig:detail}c.

Recovering underlying signal with pure black and white interphase pattern from high-level s\&p noise contamination is a very difficult problem because both signal and noise are almost binary in each channel. The method may have a hard time to distinguish which pixel is contaminated. By comparing the results shown in Figure \ref{fig:detail}f (noise2noise) and Figure \ref{fig:detail}g (ours), one may observe that our method produces higher quality images.

Noise-like patterns are common to many nature scene images, for example, the grass and the feathers of the owl in Figure \ref{fig:detail}l and the leaves in the bridge image in Table \ref{tab:pnsr_classic}. We observe that the image still looks noisy after being processed by noise2noise method, where apparent small blue and red dots stand out on the grass (Figure \ref{fig:detail}j). However, such dots are invisible in our result, as shown in Figure \ref{fig:detail}k.

\section{Conclusion} \label{sec:conclusion}

In this paper, we show that incorporating the median filtering technique in the deep neural network helps achieving compelling results in denoising the s\&p noise, especially when the noise level is high. The ability of the median layer to denoise is also experimentally testified with increasing PSNR. Our work opens the door in adopting traditional low-level nonlinear signal processing techniques in deep neural networks. The methodology of inserting non-linear spatial layers may boost the performances of some well-known deep networks.

The median is the optimum point of a set of values under $L_1$ norm, which minimizes the sum of absolute deviations. This fact makes median layers act as a regularizer to the feature channels. Unlike the annealing procedure on the loss function adopted in \cite{Lehtinen2018}, where the speed of evolving the loss from $L_2$ to $L_0$ must be carefully chosen to achieve the best result (with respect to the amount of noises), median layers is a more feasible way to control the quality of the extracted features. A single model can be trained to recover latent images with different levels of noise contaminations only using $L_2$ loss.

Spatial filtering have been invented and could be leveraged into convolutional neural networks to deal with images affected by non-linear noise. More study on the median placements could result in understanding its impact in the process.

\ifCLASSOPTIONcaptionsoff
  \newpage
\fi



\bibliographystyle{IEEEtran}
\bibliography{denoise}

\begin{thebibliography}{10}
\providecommand{\url}[1]{#1}
\csname url@samestyle\endcsname
\providecommand{\newblock}{\relax}
\providecommand{\bibinfo}[2]{#2}
\providecommand{\BIBentrySTDinterwordspacing}{\spaceskip=0pt\relax}
\providecommand{\BIBentryALTinterwordstretchfactor}{4}
\providecommand{\BIBentryALTinterwordspacing}{\spaceskip=\fontdimen2\font plus
\BIBentryALTinterwordstretchfactor\fontdimen3\font minus
  \fontdimen4\font\relax}
\providecommand{\BIBforeignlanguage}[2]{{%
\expandafter\ifx\csname l@#1\endcsname\relax
\typeout{** WARNING: IEEEtran.bst: No hyphenation pattern has been}%
\typeout{** loaded for the language `#1'. Using the pattern for}%
\typeout{** the default language instead.}%
\else
\language=\csname l@#1\endcsname
\fi
#2}}
\providecommand{\BIBdecl}{\relax}
\BIBdecl

\bibitem{zhang2017beyond}
K.~Zhang, W.~Zuo, Y.~Chen, D.~Meng, and L.~Zhang, ``Beyond a {Gaussian}
  denoiser: Residual learning of deep {CNN} for image denoising,'' \emph{IEEE
  Transactions on Image Processing}, vol.~26, no.~7, pp. 3142--3155, 2017.

\bibitem{zhang2018ffdnet}
K.~Zhang, W.~Zuo, and L.~Zhang, ``Ffdnet: Toward a fast and flexible solution
  for {CNN} based image denoising,'' \emph{IEEE Transactions on Image
  Processing}, 2018.

\bibitem{Liu2018}
\BIBentryALTinterwordspacing
D.~Liu, B.~Wen, X.~Liu, Z.~Wang, and T.~Huang, ``When image denoising meets
  high-level vision tasks: A deep learning approach,'' in \emph{Proceedings of
  the Twenty-Seventh International Joint Conference on Artificial Intelligence,
  {IJCAI-18}}.\hskip 1em plus 0.5em minus 0.4em\relax International Joint
  Conferences on Artificial Intelligence Organization, 7 2018, pp. 842--848.
  [Online]. Available: \url{https://doi.org/10.24963/ijcai.2018/117}
\BIBentrySTDinterwordspacing

\bibitem{Fu2018Multimedia}
B.~Fu, X.~Zhao, Y.~Li, X.~Wang, and Y.~Reng, ``A convolutional neural networks
  denoising approach for salt and pepper noise,'' \emph{Multimedia Tools and
  Applications}, pp. 1--18, 2018.

\bibitem{Lehtinen2018}
J.~Lehtinen, J.~Munkberg, J.~Hasselgren, S.~Laine, T.~Karras, M.~Aittala, and
  T.~Aila, ``Noise2noise: Learning image restoration without clean data,'' in
  \emph{International Conference on Machine Learning (ICML) 2018}, 2018, pp.
  2971--2980.

\bibitem{aaai_furuta_2019}
R.~Furuta, N.~Inoue, and T.~Yamasaki, ``Fully convolutional network with
  multi-step reinforcement learning for image processing,'' in \emph{AAAI
  Conference on Artificial Intelligence (AAAI)}, 2019.

\bibitem{chen2018DeepBoosting}
C.~Chen, Z.~Xiong, X.~Tian, and F.~Wu, ``Deep boosting for image denoising,''
  in \emph{European Conference on Computer Vision 2018 (ECCV)}, 2018.

\bibitem{Wang2011}
W.~Wang and P.~Lu, ``An efficient switching median filter based on local
  outlier factor,'' \emph{IEEE Signal Processing Letters}, vol.~18, pp.
  551--554, 2011.

\bibitem{Varghese2015}
J.~Varghese, N.~Tairan, and S.~Subash, ``Adaptive switching non-local filter
  for the restoration of salt and pepper impulse-corrupted digital images,''
  \emph{Arabian Journal for Science \& Engineering}, vol.~40, pp. 3233--3246,
  2015.

\bibitem{Julie2016}
J.~Delon, A.~Desolneux, and T.~Guillemot, ``Parigi: a patch-based approach to
  remove impulse-gaussian noise from images,'' \emph{Image Process On Line},
  vol.~5, pp. 130--154, 2016.

\bibitem{Huang1979}
T.~S. Huang, G.~J. Yang, and G.~Y. Tang, ``A fast two-dimensional median
  filtering algorithm,'' \emph{IEEE Transactions on Acoustics, Speech, and
  Signal Processing}, vol.~27, pp. 13--18, 1979.

\bibitem{Dong2015PAMI}
C.~Dong, C.~C. Loy, K.~He, and X.~Tang, ``Image super-resolution using deep
  convolutional networks,'' \emph{IEEE Transactions on Pattern Analysis and
  Machine Intelligence (TPAMI)}, vol.~38, pp. 295--307, 2015.

\bibitem{MartinFTM01}
D.~Martin, C.~Fowlkes, D.~Tal, and J.~Malik, ``A database of human segmented
  natural images and its application to evaluating segmentation algorithms and
  measuring ecological statistics,'' in \emph{Proc. 8th Int'l Conf. Computer
  Vision}, vol.~2, July 2001, pp. 416--423.

\bibitem{Srinivasan2007SPL}
K.~S. Srinivasan and D.~Ebenezer, ``A new fast and efficient decision-based
  algorithm for removal of high-density impulse noises,'' \emph{IEEE Signal
  Processing Letters}, vol.~14, pp. 189--192, 2007.

\bibitem{Burger2012}
H.~C. Burger, C.~J. Schuler, and S.~Harmeling, ``Image denoising: Can plain
  neural networks compete with bm3d?'' in \emph{Proc. 2012 IEEE Conference on
  Computer Vision and Pattern Recognition}, vol. 157, 2012, pp. 2392--2399.

\end{thebibliography}

%

%
%
%


\begin{IEEEbiography}[{\includegraphics[width=1in,height=1.25in,clip,keepaspectratio]{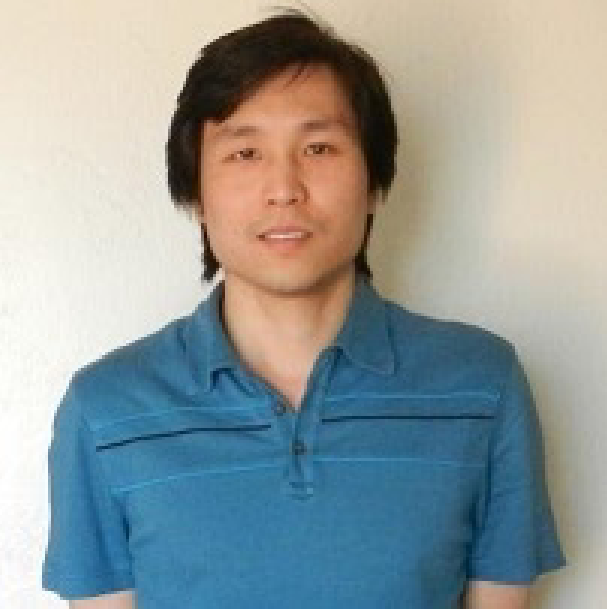}}]{Luming Liang}
is a senior researcher at Microsoft. Before working at Microsoft, Luming was a software engineer and then a data \& applied scientist at Uber and Microsoft, respectively. He received his BSc. and MSc. in the School of Information Science and Engineering from Central South University, China, in 2005 and 2008, respectively and his PhD in the Department of Electrical Engineering and Computer Science from Colorado School of Mines, USA in 2014. His primary research interest is finding shape correspondences and image analysis.
\end{IEEEbiography}

\begin{IEEEbiography}[{\includegraphics[width=1in,height=1.25in,clip,keepaspectratio]{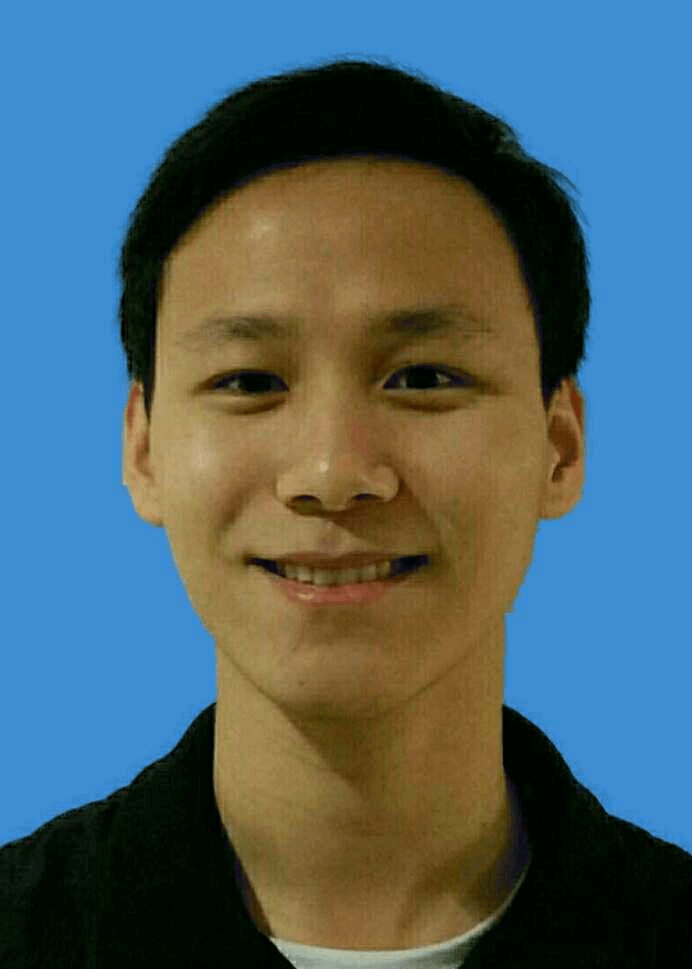}}]{Sen Deng}
is a master candidate at Nanjing University of Aeronautics and Astronautics (NUAA), China. He received his Bachelor's degree in the University of Electronic Science and Technology of China. His research interests include deep learning, image processing and computer vision.
\end{IEEEbiography}

\begin{IEEEbiography}[{\includegraphics[width=1in,height=1.25in,clip,keepaspectratio]{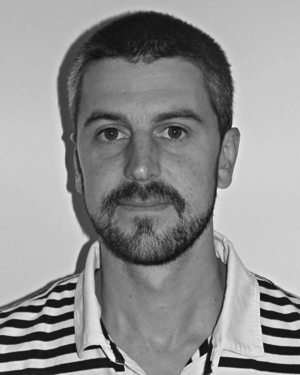}}]{Lionel Gueguen}
 is a Senior Engineer at Uber, CO, USA since 2016, where he conducts research and engineering for information extraction from images. He had been an R\&D Scientist with Image Mining Labs of DigitalGlobe Inc. Lionel received the engineering degree in telecommunications and the M.S. degree in signal and image processing from the Ecole Nationale Superieure des Telecommunications de Bretagne, Brest, France, in 2004 and the Ph.D. degree in signal and image processing from the Ecole Nationale Superieure des Telecommunications, Paris, France, in 2007.
\end{IEEEbiography}

\begin{IEEEbiography}[{\includegraphics[width=1in,height=1.25in,clip,keepaspectratio]{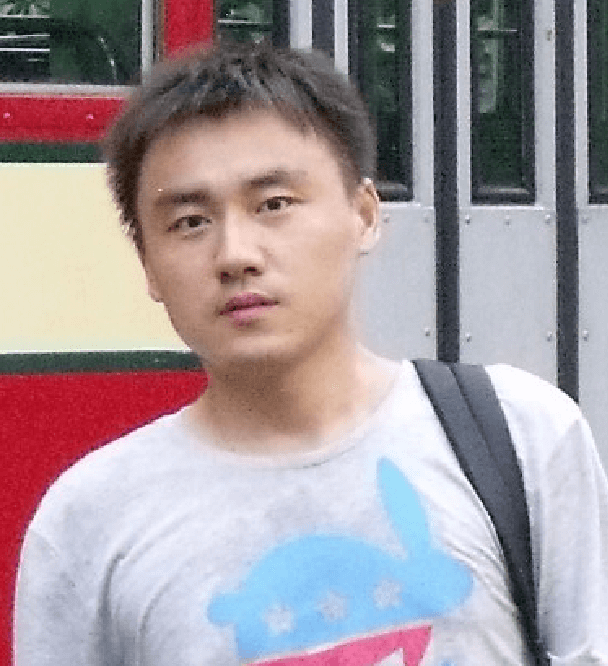}}]{Mingqiang Wei}
 received his Ph.D degree (2014) in Computer Science and Engineering from the Chinese University of Hong Kong (CUHK). He is an associate professor at the School of Computer Science and Technology, Nanjing University of Aeronautics and Astronautics (NUAA). Before joining NUAA, he served as an assistant professor at Hefei University of Technology, and a postdoctoral fellow at CUHK. His research interest is computer graphics with an emphasis on smart geometry processing.
\end{IEEEbiography}

\begin{IEEEbiography}[{\includegraphics[width=1in,height=1.25in,clip,keepaspectratio]{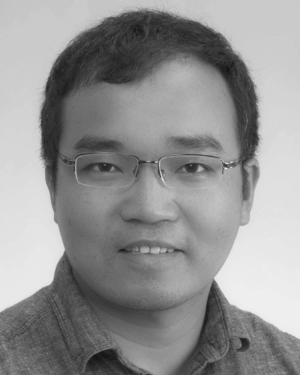}}]{Xinming Wu}
 received the Ph.D. degree in geophysics from the Colorado School of Mines, Golden, CO, USA, in 2016.  He was a member of the Center for Wave Phenomena, Colorado School of Mines. He is currently a professor at School of earth and space sciences of University of Science and Technology China (USTC). Post-Doctoral Fellow with The University of Texas at Austin, Austin, TX, USA. His research interests include image processing, 3-D seismic interpretation, subsurface modeling, and geophysical inversion.  Dr. Wu received the awards for the Best Paper in Geophysics in 2016 and the Best Student Poster Paper presented at the 2017 SEG Annual Meeting.
\end{IEEEbiography}


\begin{IEEEbiography}[{\includegraphics[width=1in,height=1.25in,clip,keepaspectratio]{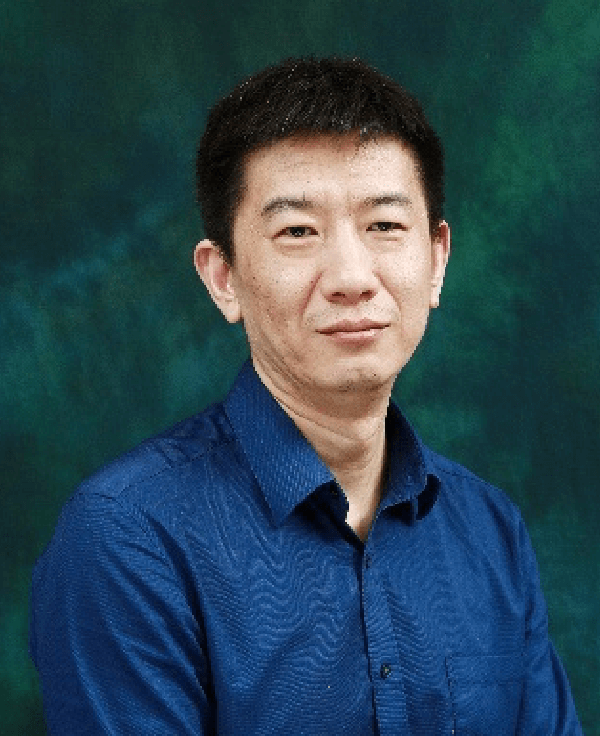}}]{Jing Qin}
is currently an assistant professor in the Center for Smart Health, School of Nursing, The Hong Kong Polytechnic University. He received his PhD degree in Computer Science and Engineering from the Chinese University of Hong Kong in 2009. Dr Qin received the Champion in the 3rd Hong Kong Innovation Day and Innovation Awards Competition, Medical Image Analysis-MICCAI'17 Best Paper Award, the Best Paper Award in Medical Image Computing in International Conference on Medical Imaging and Augmented Reality 2016, the Hong Kong Medical and Health Device Industries Association Student Research Award in 2009 and Global Scholarship Program for Research Excellence (CNOOC Grants) from CUHK in 2008. He and his collaborators were nominated for outstanding paper award in International Simulation and Gaming Association 40th Annual Conference in 2009. Dr Qin's research interests include medical image processing, virtual/augmented reality for healthcare and medicine training, deep learning, visualization and human-computer interaction and health informatics.
\end{IEEEbiography}

\end{document}